# Functional Connectivity Based Classification of ADHD Using Different Atlases

Sartaj Ahmed Salman
School of Computer Science and Engineering
Nanjing University of Science and Technology
Nanjing, China
sartajbalti@njust.edu.cn

Zhi Chao Lian
School of Computer Science and Engineering
Nanjing University of Science and Technology
Nanjing, China
izcts@163.com

Marva Saleem
School of Computer Science and Engineering
Nanjing University of Science and Technology
Nanjing, China
marvamughal@hotmail.com

Yuduo Zhang
School of Computer Science and Engineering
Nanjing University of Science and Technology
Nanjing, China
zyd@njust.edu.cn

*Abstract*—These days, the diagnosis of disorders related to neuropsychiatric by computational strategies is gaining attention day by day. It's really critical to find out the functional connectivity of brain based on Functional-Magnetic-Resonance-Imaging(fMRI) to diagnose the disorder. Till now it's known as a chronic disease and millions of children amass the symptoms of this disease, so there is a lot of vacuum for the researcher to formulate a model to improve the accuracy to diagnose ADHD accurately. In this paper we consider Functional Connectivity's of a brain extracted using various time templates/Atlases, Local-Binary Encoding-Method(LBEM) algorithm is utilized for feature extraction while Hierarchical-Extreme-Learning-Machine(HELM) is used to classify the extracted features. To validate our approach, fMRI data of 143 normal and 100 ADHD affected children is used for experimental purpose Our experimental results are based on a comparison of various Atlases given as CC400, CC200, and AAL. Our model achieves high performance with CC400 as compared to other Atlases.

## I. INTRODUCTION

The brain is the essential organ to command and correlate the actions and reactions of the human body. It is essential because it permits us to think, feel, and perform all the things that make us benevolent. Now a day's research on brain diseases is attaining so much attention and it's became a challenge for the researchers to make sure their involvement. In order to make sure that brain science is attaining courtesy, some niceties that are provided in this work. Now let's talk about a neuropsychiatric disease which is very common now a day's in children's, almost 5 % of children's of the whole world are affected [1]. This disease is known as ADHD (Attention. Deficit. hyperactivity disorder). There are three types of ADHD, named as inattentive, hyperactive-impulsive and combined. It's difficult to diagnose because it doesn't have any physical symptoms. ADHD is an emergent disease as lack of knowledge. Current research shows that this disease is not limited to children's but also surplus even in their adulthood.

It is nearly 60% that the disease surplus in their adulthood [2][3]. In 2008 G. Polanski, and P. Jensen explain that almost 5 to 6 % of school-age children's and 2 to 4 % of adults suffer from this disease [4].

It is clinched that this disease is a hot topic for the researchers of this era to develop the computational classification model of ADHD. A lot of work has been reported on automatic diagnosis methods in the literature for extracting a mass features from fMRI. These extracted features are further arranged into voxel and region level features. In 2004 Y. Zhang et al. introduced ReHo (Regional Homogeneity) [5]. ALFF (Amplitude of Low Frequency Fluctuations) was introduced by Y. Zang in 2007 and H. Yang in 2011 [6][7] explained the abnormal activities of ADHD. D. Long et al. in [5] used the extracted ReHo (Regional Homogeneity) and ALFF features from fMRI data and deployed these features to classify initial Parkinson's disease in 2014[8]. Although these voxel and region level features are so simple and instinct to calculate thus we should need to consider appropriate feature extraction methods before classification. In the present era of computer vision, the diagnosis of diseases regarding human brain is mostly based on the fMRI images. So there is an important point that how to preprocess these fMRI images. ML (Machine Learning) is receiving much attention, how to design algorithm to extract valuable features from fMRI images. Spatial and temporal feature extraction technique is also used for fMRI images in [9]. D. Liu proposed a functional-anatomical discriminative model to identify the ADHD patients in which he takes a grouped of functional connectivity networks and performed ICA (Independent Component Analysis) for feature extraction in 2010 [10].

In the past few years, a lot of automatic diagnosis methods were proposed for ADHD by various researchers [11][12][13]. They introduced various machine learning approaches in which functional connectivity's were considering as a feature of ADHD. In 2014, A. Tabas proposed a method Spatial Discriminative

Independent Component Analysis (SDICA) for brain functional networks uniting with the Linear Discriminative Analysis (LDA) [11]. There are many standpoints regarding functional connectivity, so keeping in view all these standpoints, it is clear that FC's are treated as a topo-graphic map. A lot of graph networks are also used regarding FC's for the classification of ADHD. S. Dey reported a graph-based model for the classification of ADHD using FC's in 2015, defined the edge weights with a correlation filter and model the brain FC's networks into a graph, take the nodes of the graph as an attributes/parameter and exploit it to make a graph vector distance[12]. In 2012 Che-Wei Chang et al., explained a method in which textures of anatomical regions with MRI brain data are used for the classification of ADHD[13]. In 2016 Atif riaz et al designed model for the fusion of fMRI and non fMRI data for the classification of ADHD[14]. Yan Zhang et at. presented another method for feature space separation with sparse representation. They introduced a new frame work with a dual diagnosis model based on sparse for the classification[15]. In 2018 J Lopez Marconi et al. proposed a universal background model for the classification of ADHD, it's an EEG based multi-channel and as a feature based auto-regressive model parameters [16]. Reza yaghoobi et al. in 2017 developed a mixture of an expert fuzzy model for the classification of ADHD. He also mentioned that it just not a novelty in the fuzzy theory, it's also deliberated as an improved substitute for the identification of children's with ADHD [17].

Classification rules of ADHD and NC (Normal Controls) are not easy to establish [18]. It is obligatory for understanding and deducing neurophysiology and for the study of neurological diseases that we should know about the relationship between functional and structural brain connections. From the literature, many statistical tools that have been used to measure these relationships quantitatively[19][20][21]. Various time templates were used to extract the functional brain regions, i.e. CC400, CC200 and AAL. These ROI atlases are derived by parcellating the resting state data functionally. In 2017 Lirong tan et introduced a computational model for the classification of ADHD, they used the CC400 time template for the extraction of Functional volumes, and find 10 extra regions of the brain which are being affected by ADHD. The regions were located mainly in the temporal lobe, parietal lobe, frontal lobe and occipital lobe[22]. Which helped researchers to differentiate the ADHD from NC. The volumes of the brain region are considered to be an anatomical feature, usually calculated based on anatomical magnetic resonance images. The decreasing number of anatomical volumes in a brain region is a clear example of ADHD[23].

In this paper, DPABI (Data Processing and Analysis of Brain Imaging) tool is used for the extraction and analysis of ROI's and FC's. The templates used for the extraction and analysis are CC400. it extracts 392 FC's. AAL extracts 116 FC's and CC200 extracts 200 FC's. We used LBEM for feature extraction and HELM for the classification. The major contribution of this paper is that to pre-process raw data, collected from the Peeking University. Different atlases for the extraction of functional connectivity and compare these atlases.

This paper is arranged as follows. Section 2. comprises methodology, data collection, all the pre-processing steps and FC analysis of a pre-processed data. Section 3. briefly explained the proposed algorithms. Section 4. Throws light on experimental results and analysis finally section 5 describes the conclusion and future development and further consequences.

## II. METHODOLOGY

There are four parts of our methodology as shown in Fig. 1. First, raw data was collected about ADHD affected and Normal controls. The pre-processing step has been adopted to refine the data before feature extraction, in the final step classification algorithm has been deployed for classification purpose. The above steps are discussed below one by one.

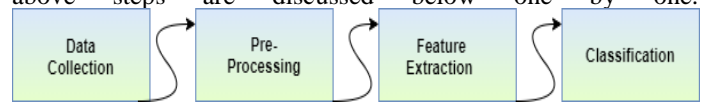

**Fig. 1.** Methodology flow diagram

### A. Data Collection

The raw data has been collected from the Peeking University of P.R China. This data related to ADHD-200 from ADHD sample home[23]. The data is publically available for research purposes to help researchers. There are 259 data samples contain 146 samples of Normal-Controls ,46 of combined-ADHD, 66 of inattentive-ADHD, and 1 of hyperactive-ADHD, each participant included in this sample consist of High resolution T1-weighted anatomical images (TI/TR/TE=1100/2530/3.25ms; flip angle=7 degrees; acquisition voxel size. = 1.3mm*1.0mm*1.3mm; scanning time=8:07min) and 176 consecutive whole brain functional volumes (TR/TE=2000/15ms; flip angle=90 degrees; acquisition-voxel size=3*3 degrees. Acquisition time = 6:00min) were obtained by using Siemens-magnetron-fast-panel-synthesizer MR2004A. Detailed information can be found on the ADHD-200-Webpage[24].

### B. Pre-Processing

To pre-process anatomical images raw data, we utilize DPABI tool[25]. There are seven steps in pre-processing followed in DPABI, which are shown in the below flow diagram Fig. 2.

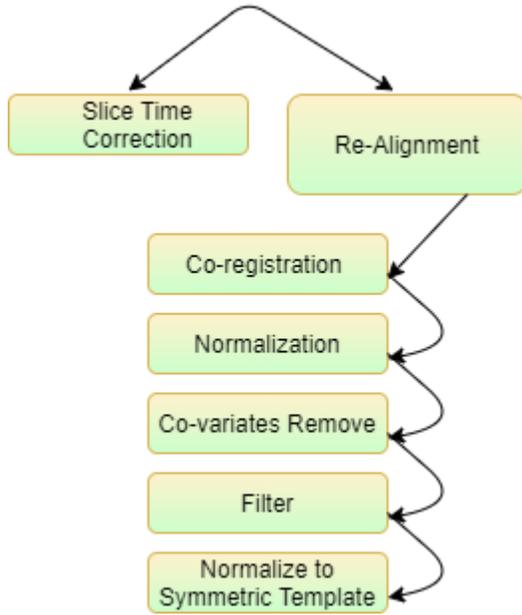

**Fig. 2.** Pre-processing Flow Diagram

Different subjects contain different time-points so we removed some time-points from each subject and make all the subjects equal. For example; Some subjects contain 235 and some contain 236-time points, we removed the first 5,6 time-points from each subject. The detail pre-processing Athena pipe-line steps as follow:

i. **Slice Time Correction** (STC) is the first step of raw fMRI data, STC is also known as interpolation, to corrects the time difference between the slices.

ii. **Realignment** is used to correct the head motion of the participant, when a participant is scanned for minutes, it human nature that it would show some movements to correct the head motion of the participant.

iii. **Co-registration** step is used as to register fMRI images into signal weighted images.

iv. **Normalization** step is adopted after Co-registration to normalize the data, it is important in brain imaging to modulate the size of the brain, normalization step is used for the modulation of brains having different size and shape into same size and shape.

v. **Co-variate** removes the nuisance covariates from the white matter regions to make the signals clearer.

vi. Regress out the motion time courses, cerebrospinal-fluid and white matter.

vii. **Band Pass filters** that has been applied having value 0.009 Hz < f < 0.08 Hz. Gaussian filter is used for spatial smoothing.

viii. In the final step, we applied the Symmetric template to sort the data.

After using the above steps, we loss some subjects from our raw data. The data we downloaded is in zip form and some subjects having problem with unzipping, also functional images are missing in some subjects. The exact data used in this paper are as follow: Normal Subjects 143, Combined 38, Hyperactive 1 and inattentive 62. In total we loss 15 subjects and the total size of sample is 244.

### C. Fucntional connectivity's

For the FC analysis, various time templates/atlases were used, which are CC400, CC200 and AAL. Each template has its own properties extracts different ROI-FC-maps. From CC400 392, CC200 200 and AAL 116 ROI-FC maps were extracted respectively. These atlases were derived and constructed by a specific scan co-relation matrix known as seed based-correlation matrix. Seed based correlation analysis (SCA) is also one of the most frequently used method to study the functional connectivity of brain. Based on the time series of seed voxels, connectivity is calculated as the time series correlation of all other voxels in the brain. The result of SCA is a map of connectivity that shows the Z-score of each voxel, indicating the correlation between its time series and the seed's time series. Each template has a different number of regions, SCA was then calculated using the subsequent BOLD signals from all possible pairs of the 392 CC400, 200 CC200 and 116 AAL.

Here Fig. 3 and Fig. 4 shows the atlas visualization and connectomes and the figure source is [23].

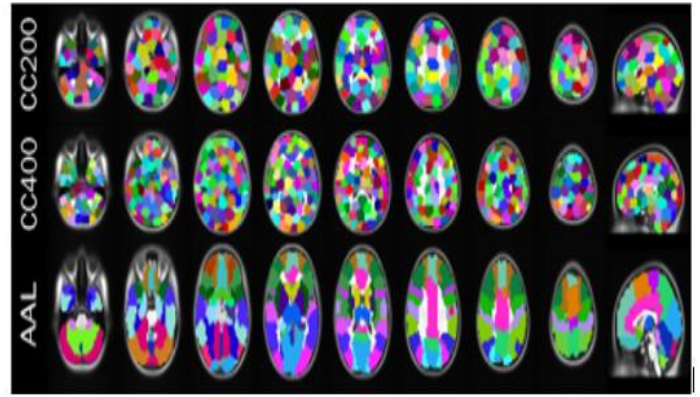

**Fig. 3.** Atlases visualization

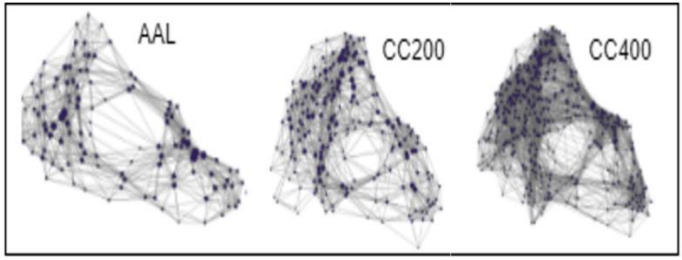

**Fig. 4.** Atlases connectomes

In the above Fig. 4 all the pre-processed data is used for the FC analysis with seed based correlation analysis and ROI-FC maps are extracted from above templets.

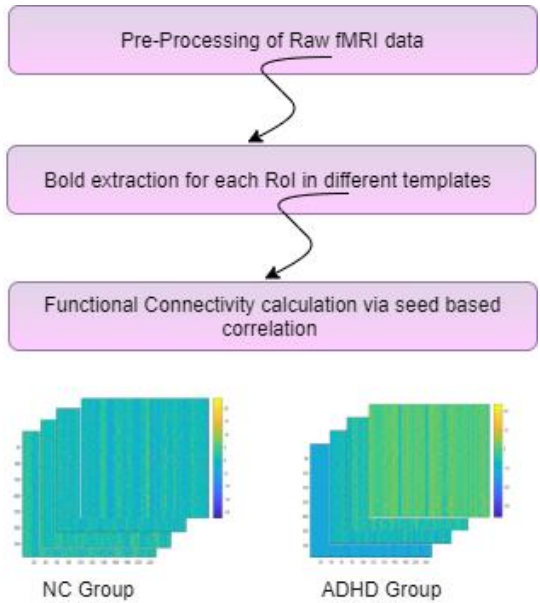

**Fig. 5.** Overall flow diagram of FC Analysis

Fig. 5 show a detail process which is adopted during functional connectivity calculation through seed based correlation method for extraction ROI-FC maps.

### III. PROPOSED ALGORITHMS

#### A. Local Binary Encoding Method

Local binary encoding short for the LBEM which is used to extract the functional interaction pattern (FIP) of the brain. our previous research in [26] proposed LBEM for the extraction of local features of fMRI data.
In this paper, LBEM is adopted as a feature extraction algorithm. We used the processed-data as a matrix which is represented below: $P = (P_1, P_2, P_3, \ldots, P_N) \in F^{M \times N}$ Where $N$ represent number of time points in each subject while $M$ is the number of regions whose FC's are extracted. The value of $F^{M \times N}$ is different for each template/atlas, $F^{392 \times N}$ for CC400, $F^{200 \times N}$ for CC200 and $F^{116 \times N}$ for AAL respectively. Here for each vector $P_n = (p_1, p_2, p_3, \ldots, p_m)$, $1 \leq n \leq N$, the number of FC's in each time-point is represented by N. We re-code the vector column $P_n$ one by one in a data matrix $P$. After re-coding, we compared the elements in vector $P_n$ with the elements in the neighbor and build a new vector $V_n = (v_1, v_2, v_3, \ldots, v_x)$, $1 \leq n \leq N$ here x value would vary template to template e.g. 782 for CC400, 398 for CC200, and 230 for AAL. A new build vector $V_n$ is calculated with the help of below equations (1-3):

$$v_{2(i-1)-1} = \begin{cases} 1, p_i \leq p_{i-1} \\ 0, p_i > p_{i-1} \end{cases}, 2 \leq i \leq M \quad (1)$$

$$v_{2(i-1)} = \begin{cases} 1, p_i \leq p_{i+1} \\ 0, p_i > p_{i+1} \end{cases}, 2 \leq i \leq M-1 \quad (2)$$

IV.
$$v_x = \begin{cases} 1, p_M \leq p_i \\ 0, p_M > p_i \end{cases} \quad (3)$$

By these equations, we got the binary form vector containing the value of 0 and 1 elements $V_n = (V_1, V_2, V_3, \ldots, V_x) \in F^{x \times N}$. The process of Encoding into Binary and Binary to Decimal is shown in Fig. 6. As shown in figure we have vector $A = (a_1, a_2, a_3, a_4)$, is encoded into binary to get a new Binary vector $B = (b_1, b_2, \ldots, b_6)$. Whenvere of $a_2 < a_1$ in vector A, it put 1 in vector B to replace $b_1$, and when $a_2 > a_3$ it assigns 0 to $b_2$ and the process of encoding into a binary would continue until we get the remaining element of vector B.. .In the next step, we converted all binary values of vector B into decimal. , All six elements in vector B is divided into different groups, each of the binary form groups is converted into decimal form within the range $[0,63]$. After transforming $B$, we got a new data matrix $Z = (Z_1, Z_2, Z_3, \ldots, Z_N) \in F^{y \times N}$, here the value of $y$ is vary template to template i.e, for CC400 $y = 112$, this data matrix represents the Local information between the FC's and it's adjacent FC's.

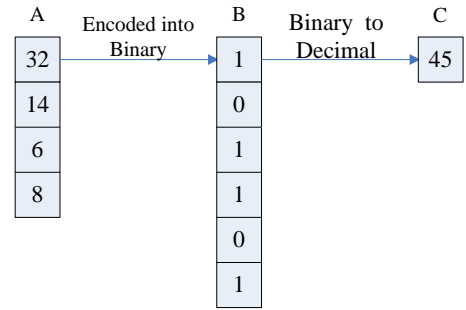

**Fig. 6.** A is encoding the vector into Binary Vector B and convert Binary Vector B into Decimal C.

## B. Extreme Learning Machine

For the detail and better understanding let's briefly review the ELM theory in [27]. ELM has only one hidden layer H with L hidden nodes. For g-classification problem, a training set is given $\{(x_j, z_j) | x_j \in R^d, z_j \in R^G\}$, where $x_j$ is a training vector, $z_j$ is a training label, and $j = 1, 2, \cdots, N$. The ELM algorithm can be described as follow:

Calculate the hidden layer matrix H

$$H = \begin{bmatrix} h(x_1) \\ \vdots \\ h(x_N) \end{bmatrix} \begin{bmatrix} h(w_1 x_1 + b_1) & \cdots & h(w_L x_1 + b_L) \\ \vdots & \ddots & \vdots \\ h(w_1 x_N + b_1) & \cdots & h(w_L x_N + b_L) \end{bmatrix} \quad (4)$$

where $h(\cdot)$ denotes the nonlinear activation function, the input weights $w_i$ and biases $w_i$ are randomly generated, I =1, 2…, L. The output weight vector β can be calculate by equation (5).

$$\beta = H^\dagger Z \quad (5)$$

where $H^\dagger$ is the Moore-Penrose generalized inverse of matrix H, and Z is the training label matrix

$$Z = \begin{bmatrix} z_1^T \\ \vdots \\ z_N^T \end{bmatrix} = \begin{bmatrix} z_{11} & \cdots & z_{1G} \\ \vdots & \ddots & \vdots \\ z_{N1} & \cdots & z_{NG} \end{bmatrix} \quad (6)$$

The corresponding result of ELM can be obtained as follow:

$$f(x_j) = h(x_j)\beta \quad (7)$$

## C. Hierarchical Extreme Learning Machine

The structure of HELM [28] is more complex as compared to ELM. The training process of HELM is different from the greedy lay wise [29] and the structure of the HELM is shown in Fig. 7. It consists of two parts: (1) unsupervised hierarchical feature selection based on ELM sparse self-coding, (2) supervised classification based on ELM.

For the G classification problem, given a training set $\{(x_j, z_j) | x_j \in R^d, z_j \in R^G\}$, where $x_j$ is the training vector, $z_j$ is the training label, and $j = 1, 2, \cdots, N$. Before unsupervised hierarchical feature extraction, the dataset should be normalized to [0, 1]. The label is resized to the G dimension, which is - 1 or 1. Input data is mapped to the ELM random feature space. We extract the features through n-layers unsupervised learning. For the HELM learning algorithm, the input layer is 0. The first layer is the hidden layer. For first layer weight $\beta_1$ the ELM sparse auto-encoder shown in Fig. 4 (a) is used to learn, and other hidden layer weights are also learned by ELM sparse auto-encoder. The results of each hidden layer can be obtained through:

$$H_i = g(H_{i-1} \cdot \beta^T), 1 \leq i \leq n \quad (8)$$

where $H_i$ is the output of $i^{th}$ layer and $H_{i-1}$ is the output of the $(i-1)^{th}$ layer, $g(.)$ is the activation function of the hidden layers, and $\beta$ is the $i^{th}$ hidden layer weight (output weight) [30]. For HELM, each hidden layer is a separate module. Once the features of the previously hidden layer are extracted, the weight of the current hidden layer would be fixed without fine-tuning [31].

Using the L1 norm regularization method to paradigm ELM auto-encoder, the ELM sparse auto-encoder has enhanced parameter generalization performance. Besides, the use of penalty terms can constrain the characteristics of our model. This made the learning model sparser. The ELM sparse auto-encoder equation of the optimization model is described in the below formula.:

$$O_\beta = argmin_\beta \{||A\beta - X||^2 + ||\beta||_{l1}\} \quad (9)$$

Where A is the random mapping output, $\beta$ is the weight of the hidden layer, and X represents the input data. In the proposed automatic encoder, A is a random initialization output mapped by a random weight matrix $b = [b_1, b_2, \ldots, b_n]$, required no optimization.. The structure of ELM sparse automatic encoder is shown in Fig. 7 (a) (b). It improves the accuracy and computational time as well.

In this phase of the paper, we described the ELM optimization algorithm based on sparse self-coding. Fast iterative shrinkage threshold algorithm (FISTA) is used to solve the constrained minimization problem of unceasing differentiable functions. We used the constant step type of FISTA. The precise algorithm is described below:

I. Calculate the Lipschitz constant $\gamma$ of the gradient $\nabla p$ of the function $||A\beta - X||^2$

II. Calculate β iteratively. Begin the reiteration by taking $y_1 = \beta_0 \in R^n$, t1=1 as the initial points. Then, for i (i≥ 1) S

**Step 1:** $\beta_i = S_\gamma(y_i)$, Where

$$S_\gamma = arg\, min_\beta \{\frac{\gamma}{2} ||\beta - (\beta_{i-1} - \frac{1}{\gamma}(\beta_{i-1}))||^2 + ||\beta||_{l1}\}$$

**Step 2:** $t_{i+1} = \frac{1+\sqrt{1+4t_i^2}}{2}$

**Step 3:** $y_{i+1} = \beta_i + \left(\frac{t_{i-1}}{t_{i+1}}\right)(\beta_i - \beta_{i-1})$

The second part is supervised feature classification based on ELM. Elm input $H_n$ is the output matrix of n-layer ELM sparse auto encoder.

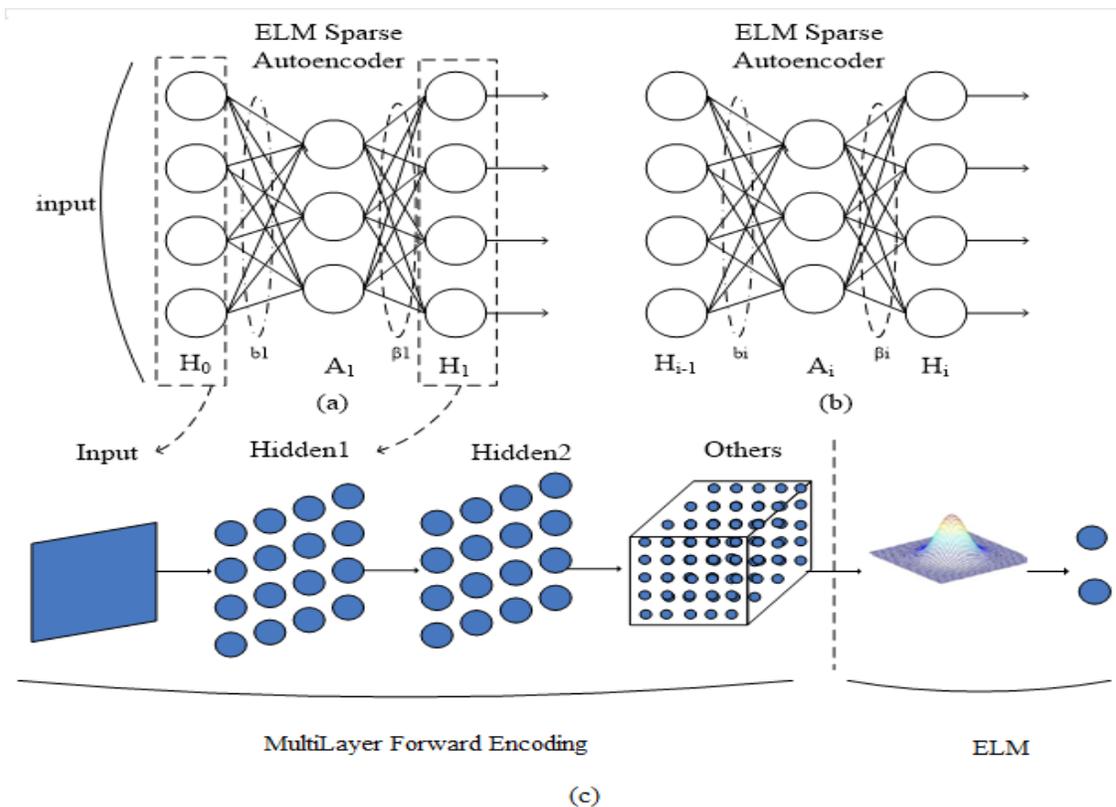

**Fig. 7.** a) ELM Auto Sparse Encoder structure of 1st hidden layer. (b) ELM Sparse Encoder structure of Ith hidden layers. (c) The overall architecture of HELM [32]

## IV. EXPERIMENTAL RESULTS AND ANALYSIS

In this section, to prove our method's accuracy and proficiency for the classification purpose, we performed repetitive experiments on our ADHD r-fMRI dataset which contains 143 typically growing children's and 100 ADHD effected children's. we selected different regions of the brain using different time templates i.e. CC400, CC200, and AAL followed by 392, 200, and 119 regions. These regions are further used as a node of a network correspondingly. These different regions for all 243 subjects were determined using the common tool DPABI [25]. After pre-processing, we get different time series matrix for different atlases for every sample, 392*230 for CC400, 200*230 for CC200, and 116*230 for AAL respectively. And then the r-fMRI time series of each sample is used for the additional experiment.

In this work, the 5-fold cross-validation method is used to appraise the proposed method. The experiment was repeated 30 times and then we calculate the average classification accuracy of each fold. The experiment is implemented on two different methods using different time templates (atlases) to compare to compare the accuracy, the methods are ELM and HELM while the three-time templates are as follows CC400, CC200, and AAL. First, Using ELM with the given templates, it is cleared that the CC400 achieves a better result than the rest and the classification accuracy is shown in Table 1 (average accuracy of ADHD is 96% and NC is 98%) the accuracy using CC200 and AAL is shown in table 1 below. This clearly illustrates that the accuracy of ELM using CC400 is better than CC200 and AAL.

After ELM we used HELM with different atlases and the classification accuracy is shown in Table 2 (average accuracy of ADHD is 98 % and NC is 99.5 %) the accuracy of remaining atlases is shown in Table 2 below. These experiments demonstrate that CC400 achieves a better result than other atlases and the accuracy using HELM is quite better than ELM which shows that using ELM sparse auto-encoder to extract features from training data-set and testing data-set before HELM can improve the classification accuracy.

Finally, in the last section of experiments, we compared the results of HELM using different layers, and the experiment displays a little difference in accuracy with different layers. Table 3 (average accuracy of NC is 99.72 % and ADHD is 98.06 %) is the experimental result of HELM using CC400 up to three hidden layers. Similarly, Table 4 (the average accuracy of NC is 97.08 % and ADHD is 96.56 %) is the experimental results of HELM using CC200 up to three hidden layers and Table 5 (average accuracy of NC is 94.30 % and ADHD is 92.81 %) it noticed that the precision with one hidden layer is slightly more than with the multiple layers. It illustrates that with one hidden layer the accuracy is more and reduces a little with the increasing number of layers. The reason is quite clear it is because of the data, the data used in this paper is not that large as the method needs. The experiment was also repeated with a small data set than the data we used in this paper, which clearly shows that with the increasing number of the layer the result we get is more worst. The results are quite stable with the given data and it's also clear that if the data is large as the method needs, then the classification accuracy will increase with the number of hidden layers.

## V. CONCLUSION

In this paper, we developed a dataset for experimental purposes. To execrate effective features, form our dataset, the feature extraction algorithm is designed. Classification scheme is proposed to improve the accuracy of brain data classification to differentiate ADHD patients and Normal Controls (NC). The innovation of our paper is the use of hierarchical ELM sparse auto-encoder to extract features before ELM which outperformed other classifiers that have been used to differentiate ADHD patients and NC while requiring fewer hidden layers. The comparison of experimental results at different scales shows that the classification accuracy with one hidden layer is higher, and the selection of the number of layers for classifier feature extraction is not as good as possible but the result is quite satisfying and stable. As future work, we hope to have more experiments to improve the average accuracy of fMRI classification to differentiate ADHD patients and NC (Normal Controls).
.

ACKNOWLEDGMENT

This work was supported by the Major Research plan of the National Natural Science Foundation of China (No. 91538108), the National Natural Science Foundation of China (No. 61501241,61571230), the Natural Science Foundation of Jiangsu Province (No. BK20150784, BK20150792, BK20161500), and the China Postdoctoral Science Foundation (No. 2015M570450, 2015M581800).

TABLE I.    AVERAGE ACCURACY USING ELM WITH DIFFERENT ATLASES

| Class | ELM with CC400 | | ELM with CC200 | | ELM with AAL | |
|---|---|---|---|---|---|---|
| | NC | ADHD | NC | ADHD | NC | ADHD |
| Fold 1 | 0.9747 | 0.9652 | 0.9747 | 0.9522 | 0.9231 | 0.8761 |
| Fold 2 | 0.9779 | 0.9549 | 0.9576 | 0.9449 | 0.9123 | 0.9387 |
| Fold 3 | 0.9835 | 0.9974 | 0.8924 | 0.8853 | 0.9324 | 0.8321 |
| Fold 4 | 0.9711 | 0.9564 | 0.9612 | 0.8621 | 0.9713 | 0.8762 |
| Fold 5 | 0.9752 | 0.9048 | 0.9332 | 0.9731 | 0.8951 | 0.8955 |
| Average | 97.65 % | 95.57 % | 94.38% | 92.35% | 92.68% | 88.32% |

TABLE II.    AVERAGE ACCURACY USING HELM WITH DIFFERENT ATLASES

| Class | HELM with CC400 | | HELM with CC200 | | HELM with AAL | |
|---|---|---|---|---|---|---|
| | *NC* | *ADHD* | *NC* | *ADHD* | *NC* | *ADHD* |
| Fold 1 | 0.9932 | 0.9941 | 0.9928 | 0.9762 | 0.9931 | 0.9861 |
| Fold 2 | 0.9979 | 0.9587 | 0.9593 | 0.9829 | 0.9753 | 0.9767 |
| Fold 3 | 1 | 0.9924 | 0.9286 | 0.8997 | 0.9424 | 0.8931 |
| Fold 4 | 0.9981 | 0.9956 | 0.9801 | 0.9965 | 0.9113 | 0.8892 |
| Fold 5 | 0.9972 | 0.9625 | 0.9932 | 0.9731 | 0.8932 | 0.8955 |
| Average | 99.72 % | 98.06 % | 97.08% | 96.56% | 94.30% | 92.81% |

TABLE III. AVERAGE ACCURACY USING HELM WITH CC400 AND MULTIPLE LAYERS

| Class Using CC400 | HELM with one hidden layer | | HELM with two hidden layer | | HELM with three hidden layer | |
|---|---|---|---|---|---|---|
| | *NC* | *ADHD* | *NC* | *ADHD* | *NC* | *ADHD* |
| Fold 1 | 0.9932 | 0.9941 | 0.9973 | 0.9852 | 0.9961 | 0.9936 |
| Fold 2 | 0.9979 | 0.9587 | 0.9895 | 0.9949 | 0.9813 | 0.9877 |
| Fold 3 | 1 | 0.9924 | 0.9867 | 0.9785 | 0.9424 | 0.9561 |
| Fold 4 | 0.9981 | 0.9956 | 0.9853 | 0.9665 | 0.9513 | 0.8792 |
| Fold 5 | 0.9972 | 0.9625 | 0.9925 | 0.9371 | 0.9932 | 0.9655 |
| Average | 99.72 % | 98.06 % | 99.02% | 97.24% | 97.28% | 95.64% |

TABLE IV. AVERAGE ACCURACY USING HELM WITH CC200 AND MULTIPLE LAYERS

| Class Using CC200 | HELM with one hidden layer | | HELM with two hidden layer | | HELM with three hidden layer | |
|---|---|---|---|---|---|---|
| | *NC* | *ADHD* | *NC* | *ADHD* | *NC* | *ADHD* |
| Fold 1 | 0.9928 | 0.9762 | 0.9616 | 0.9267 | 0.9491 | 0.9318 |
| Fold 2 | 0.9593 | 0.9829 | 0.9765 | 0.9346 | 0.9283 | 0.9257 |
| Fold 3 | 0.9286 | 0.8997 | 0.9437 | 0.9672 | 0.9424 | 0.9061 |
| Fold 4 | 0.9801 | 0.9965 | 0.9213 | 0.9263 | 0.9713 | 0.9742 |
| Fold 5 | 0.9932 | 0.9731 | 0.9615 | 0.9489 | 0.9182 | 0.8955 |
| Average | 97.08% | 96.56% | 95.29% | 94.17% | 94.18% | 92.66% |

TABLE V. AVERAGE ACCURACY USING HELM WITH AAL AND MULTIPLE LAYERS

| Class Using AAL | HELM with one hidden layer | | HELM with two hidden layer | | HELM with three hidden layer | |
|---|---|---|---|---|---|---|
| | *NC* | *ADHD* | *NC* | *ADHD* | *NC* | *ADHD* |
| Fold 1 | 0.9931 | 0.9861 | 0.8922 | 0.8265 | 0.8413 | 0.8378 |
| Fold 2 | 0.9753 | 0.9767 | 0.8908 | 0.8743 | 0.8783 | 0.8754 |
| Fold 3 | 0.9424 | 0.8931 | 0.9617 | 0.9972 | 0.9024 | 0.9041 |
| Fold 4 | 0.9113 | 0.8892 | 0.9633 | 0.9363 | 0.9213 | 0.8972 |
| Fold 5 | 0.8932 | 0.8955 | 0.9415 | 0.9189 | 0.9382 | 0.9245 |
| Average | 94.30% | 92.81% | 92.99% | 91.06% | 89.63% | 88.78% |